\documentclass[10pt,a4paper]{article}
\usepackage{f1000_styles}

\usepackage{multirow}
\usepackage{colortbl}
\usepackage{subcaption}
\usepackage{tabularx}
\usepackage[framemethod=TikZ]{mdframed}  

\definecolor{userColor}{HTML}{ECEFF1}
\definecolor{ourcolor}{HTML}{FFECB3}
\definecolor{ourcolorbold}{HTML}{FF6F00}
\definecolor{llamacolor}{HTML}{BBDEFB}
\definecolor{chatgpt}{HTML}{B2DFDB}

\usepackage{lipsum}

\newmdenv[
    backgroundcolor=ourcolor,
    linecolor=ourcolor,
    roundcorner=4pt,
    leftmargin=0pt,
    rightmargin=32pt,
    innerleftmargin=8pt,
    innerrightmargin=8pt,
    innertopmargin=4pt,
    innerbottommargin=4pt,
    skipabove=2pt,
    skipbelow=2pt,
    fontcolor=purple,
]{assistantbox}

\newmdenv[
    backgroundcolor=ourcolor,
    linecolor=ourcolor,
    roundcorner=4pt,
    leftmargin=0pt,
    rightmargin=32pt,
    innerleftmargin=8pt,
    innerrightmargin=8pt,
    innertopmargin=4pt,
    innerbottommargin=4pt,
    skipabove=2pt,
    skipbelow=2pt,
    fontcolor=black,
]{assistantboxdark}

\newmdenv[
    backgroundcolor=llamacolor,
    linecolor=llamacolor,
    roundcorner=4pt,
    leftmargin=0pt,
    rightmargin=32pt,
    innerleftmargin=8pt,
    innerrightmargin=8pt,
    innertopmargin=4pt,
    innerbottommargin=4pt,
    skipabove=2pt,
    skipbelow=2pt,
    fontcolor=black,
]{llamabox}

\newmdenv[
    backgroundcolor=userColor,
    linecolor=userColor,
    roundcorner=4pt,
    leftmargin=32pt,
    rightmargin=0pt,
    innerleftmargin=8pt,
    innerrightmargin=8pt,
    innertopmargin=4pt,
    innerbottommargin=4pt,
    skipabove=2pt,
    skipbelow=2pt,
    fontcolor=black,
]{userbox}

\newmdenv[
    backgroundcolor=chatgpt,
    linecolor=chatgpt,
    roundcorner=4pt,
    leftmargin=0pt,
    rightmargin=32pt,
    innerleftmargin=8pt,
    innerrightmargin=8pt,
    innertopmargin=4pt,
    innerbottommargin=4pt,
    skipabove=2pt,
    skipbelow=2pt,
    fontcolor=black,
]{chatgptbox}

\usepackage[numbers]{natbib}

\begin{document}
\pagestyle{fancy}

\title{SituationalLLM: Proactive language models with scene awareness for dynamic, contextual task guidance}
\author[1$\dagger$]{Muhammad Saif Ullah Khan}
\author[1$\dagger$]{Muhammad Zeshan Afzal}
\author[1]{Didier Stricker}
\affil[1]{German Research Center for Artificial Intelligence (DFKI), 67663 Kaiserslautern, Germany}

\maketitle
\thispagestyle{fancy}

$\dagger$ Corresponding author: \texttt{muhammad\_saif\_ullah.khan@dfki.de}
\\
\\
\begin{abstract}

\paragraph{Background.}
Large Language Models (LLMs) have demonstrated remarkable success in text-based reasoning tasks but struggle to provide actionable guidance in real-world physical environments. This limitation arises from their lack of situational awareness—an inability to recognize gaps in their understanding of a user’s physical context, leading to unreliable and overly generic instructions. To address this, we propose \textbf{SituationalLLM}, a novel approach that integrates structured scene representations into LLMs to improve context-aware assistance.

\paragraph{Methods.}
SituationalLLM leverages scene graphs—structured representations of objects, attributes, and spatial relationships—to encode real-world environments in a text-based Scene Graph Language. We introduce the \textbf{Situational Awareness Database for Instruct-Tuning (SAD-Instruct)}, which pairs diverse scene graphs with multi-agent dialogue, enabling LLMs to iteratively refine their guidance through clarifying questions. A\textbf{ LoRA-adapted LLaMA-3-8B model} is fine-tuned on SAD-Instruct to bridge structured knowledge with natural language reasoning, enhancing its ability to recognize missing information and dynamically adjust responses.

\paragraph{Results.}
Qualitative evaluations show that SituationalLLM outperforms state-of-the-art LLMs (GPT-4, LLaMA-3) in providing precise, task-specific, and contextually relevant instructions. The model reduces hallucinations by proactively identifying missing environmental details and requesting clarifications before generating guidance. Through comparative analyses on everyday tasks (e.g., cooking, office assistance), SituationalLLM demonstrates superior adaptability, delivering grounded, user-centered recommendations.

\paragraph{Conclusion.}
By integrating structured scene representations and iterative dialogue-based refinements, SituationalLLM enables more reliable, context-aware AI assistants. This research highlights the significance of bridging structured knowledge with natural language for enhanced real-world task guidance. Future work should focus on expanding scenario diversity and improving real-time scene perception to further enhance situational adaptability.
\end{abstract}    

\section*{\color{OREblue}Keywords}

Large Language Models, Scene Graphs, Context-Aware Assistance,

Situational Awareness.

\clearpage
\pagestyle{fancy}

\section{Introduction}
\label{sec:intro}

Large language models (LLMs) have demonstrated exceptional capabilities in understanding and generating human language~\cite{radford2019language,brown2020language,gpt4vmodelcard}. This has driven advancements in a wide range of tasks, ranging from language understanding~\cite{gong2024llms,wang2024recent} and software development~\cite{ozkaya2023application} to human motion descriptions~\cite{khan2024human}. Recent work is increasingly focused on creating LLM-based AI systems that operate in real-world environments~\cite{wang2024survey}. This includes both virtual assistants like chatbots and embodied agents like robots. However, traditional LLMs struggle to leverage accurate world models of their surroundings, limiting their effectiveness in physical contexts~\cite{chen2024can}. This is a significant barrier to developing robust LLM-driven systems that can help complete tasks in the real world.

Current LLMs often fail to recognize their incomplete understanding of the physical world and rarely seek clarifications~\cite{chen2024see}. Consequently, when used for task guidance, they tend to produce verbose and overly generic instructions with hallucinations that rely on broad assumptions, making them difficult for users to interpret and apply in their situation. For embodied agents, this can lead to ineffective or even unsafe task execution~\cite{zhu2024riskawarebench}, highlighting the critical need for actionable and contextually accurate task guidance.

\begin{figure}[ht]
    \centering
    \scriptsize
    \begin{subfigure}{0.51\linewidth}
        \begin{userbox}
        I need help opening this glass jar on my desk.
        \end{userbox}

        \begin{chatgptbox}
        Opening a stubborn glass jar can be tricky, but here are a few techniques that can help:
        \end{chatgptbox}

        \begin{chatgptbox}
         1. \textbf{Grip Enhancement:} \begin{itemize}
            \item Use a rubber glove or a dish towel to increase grip on the lid.
            \item Alternatively, wrap a thick rubber band around the lid for better traction.
         \end{itemize}
        \end{chatgptbox}

        \begin{chatgptbox}
         2. \textbf{Tapping the Lid:} \begin{itemize}
            \item Gently tap around the edges of the lid with the back of a knife or a wooden spoon. This may loosen the seal.
         \end{itemize}
        \end{chatgptbox}
    \end{subfigure}
    \hfill
    \begin{subfigure}{0.44\linewidth}
        \begin{chatgptbox}
         \begin{itemize}
            \item You can also tap the bottom of the jar against a soft surface to release pressure inside.
         \end{itemize}
        3. \textbf{Hot Water Trick:} \begin{itemize}
            \item Run hot water over the lid (not the jar itself) for about 30 seconds. The metal expands faster than the glass, making it easier to twist.
        \end{itemize}
        4. \textbf{Leverage Tools:} \begin{itemize}
            \item If you have a jar opener, it's perfect for this. If not, try using a spoon or a butter knife to pry a small gap under the lid to let air in.
        \end{itemize}
    
        Give one of these a try and let me know how it goes!
        \end{chatgptbox}
    \end{subfigure}
    \caption{GPT-4 provides comprehensive but generic guidance when assisting with physical tasks, failing to account for specific user situations and constraints. It presumes that the jar is "stubborn" and neglects to ask for details like the type of jar or the user's limitations, which can lead to less applicable advice. \textit{An ideal LLM-driven AI assistant should provide tailored advice, considering the user's real-world situation, implying a need for awareness of their physical context.}}
    \label{fig:chapgpt-4-bad-guidance-example}
\end{figure}

Figure~\ref{fig:chapgpt-4-bad-guidance-example} demonstrates this limitation with an example from GPT-4~\cite{gpt4vmodelcard}. Here, the model assumes the jar is stubborn and provides broad, non-specific solutions without understanding the user's exact context (e.g., why help is needed, the jar's characteristics, etc.). In contrast, if the user asks another human for help with a similar task, that person would likely attempt to understand why the user needs assistance before providing a targeted response specific to the user's circumstances. The LLM's lack of awareness about the user's context and inability to recognize its own incomplete understanding, or what we term situational awareness~\cite{endsley1995toward,linghu2024multi}, results in ill-suited guidance. This highlights the need for an LLM to recognize incomplete context and proactively seek details to refine its guidance.

The lack of detailed descriptions of the physical world in standard user interactions with LLMs makes it difficult for LLMs to understand the context. For LLM-driven virtual assistants to function more naturally and be user-friendly, instead of relying entirely on the user to provide all necessary context, a practical assistant should proactively identify missing details and ask clarifying questions, much like a human would. This ability to extract context through interaction would make systems more intuitive and reduce the burden on users to provide complete, precise information upfront. In the case of LLM-driven embodied agents, some environmental context can be gleaned through sensors attached to the agent. When combined with scene-understanding tools, these sensors can offer structured representations of the environment—scene graphs—which can then be added to prompts to improve situational awareness.

However, conventional LLMs are not inherently designed to process or reason with structured data like scene graphs. To address this gap, we propose a lightweight \textit{Scene Graph Language} that encodes objects, their attributes, and the relationships between them as text. By converting each scene’s data into a standardized textual form, the LLM can parse and reason about the environment more effectively. This format is particularly crucial in real-world settings where contextual details—such as the number of objects, their positions, or their relationships—impact task feasibility.

Despite the potential of incorporating such structured information, there is a lack of standardized methods for integrating environmental inputs into LLM workflows. This makes producing actionable, context-aware responses tailored to users’ needs challenging. In response, we introduce the \textbf{Situational Awareness Database for Instruct-Tuning (SAD-Instruct)}, a novel dataset that pairs scenario-specific scene graphs with multi-agent dialogue to generate context-sensitive instructions. SAD is derived from large-scale 3D semantic scene graphs~\cite{3DSSG2020,wald2019rio}, ensuring realistic and diverse indoor scenarios. Alongside scene-graph pruning and multi-turn dialogue, SAD includes a wide range of tasks and reflection-style interactions designed to teach LLMs how to identify context gaps and ask clarifying questions.  

To demonstrate the effectiveness of our dataset in eliciting desired behaviors in LLMs, we propose \textbf{SituationalLLM}, a model fine-tuned using LoRA~\cite{hu2021lora} to effectively handle scene graph inputs, query missing details, and provide more grounded, user-centered guidance. Our method demonstrates the importance of bridging structured representations (scene graphs) with natural language dialogues, enabling LLMs to reason about what matters in a given situation.

We present initial qualitative examples showing SituationalLLM outperforming GPT-4 and Llama 3 in situational awareness by dynamically adapting to user needs and leveraging environmental context. While these results are promising, they represent a step toward more robust, context-aware AI systems rather than a definitive solution.

\subsection{Contributions}
\begin{itemize}
    \item We propose SituationalLLM, a fine-tuned language model designed to integrate structured scene knowledge for actionable, context-aware task guidance, benefiting both virtual assistants and embodied agents.
    \item We introduce the Situational Awareness Database for Instruct-Tuning (SAD-Instruct), a novel dataset that teaches LLMs to reason over structured and unstructured knowledge and refine guidance through iterative dialogue.
    \item We demonstrate, through qualitative examples, that SituationalLLM improves task specificity, reduces hallucinations, and provides grounded, reliable assistance, setting a foundation for AI systems capable of building and leveraging world models.
\end{itemize}
\section{Methods}
\label{sec:method}

This section presents our approach for creating \textbf{SituationalLLM}, which provides actionable, context-aware task guidance by leveraging structured scene graphs and interactive multi-agent dialogue. We begin with a high-level overview of our solution, introduce the Scene Graph Language for structured representations, detail the construction of the Situational Awareness Database (SAD), and then describe how we fine-tune our model.

\subsection{Overview of the proposed approach}
\label{sec:method_overview}

Traditional LLMs struggle with executing tasks in the physical world because they rely primarily on text-based pretraining and often fail to incorporate crucial environmental details. Our proposed solution, \textbf{SituationalLLM}, addresses this challenge in three key steps:

\begin{enumerate}
    \item \textbf{Encode environments as scene graphs:} We use the 3DSSG~\cite{3DSSG2020} dataset to obtain comprehensive semantic scene graphs, capturing objects, attributes, and relationships in real-world indoor environments (Sec.~\ref{sec:scene_graph_lang}).
    \item \textbf{Construct SAD with scenario-specific dialogues:} We generate situational contexts (scenarios) from these scene graphs and employ a multi-agent system to produce iterative dialogues and step-by-step instructions (Sec.~\ref{sec:sad_construction}). 
    \item \textbf{Fine-tune an LLM with scene graph awareness:} Finally, we train a LoRA adapter~\cite{hu2021lora} on top of LLaMA-3-8b-Instruct~\cite{llama3modelcard} to integrate structured and unstructured knowledge (Sec.~\ref{sec:training_situationalllm}), yielding the \textit{SituationalLLM} model.
\end{enumerate}

\subsection{Scene graph language and representation}
\label{sec:scene_graph_lang}

Scene graphs offer structured, graph-based representations of environments, encoding objects (\textit{nodes}), their attributes, and pairwise relationships (\textit{edges}). To make such information accessible to an LLM, we propose converting it into a standardized \textbf{Scene Graph Language}:

\begin{quote}
    \texttt{obj-<label>-<id>:[<attr1>,<attr2>,\ldots]; \ldots;}
    
    \texttt{rel-<id>:(<subject>-<id>,<predicate>,<object>-<id>); \ldots;}
\end{quote}

Each \texttt{obj-<label>-<id>} entry details the object’s name, ID, and any relevant attributes (e.g., “wooden,” “large,” “on table”). Relationship lines specify how objects interact (e.g., “under,” “contains,” “next to”). We can teach standard LLMs to parse and reason over environmental contexts without specialized architectures by translating scene graphs into this uniform textual format and fine-tuning lightweight adapters.

\subsection{Constructing the situational awareness database (SAD)}
\label{sec:sad_construction}

\begin{figure*}[ht]
    \centering
    \begin{subfigure}{0.72\linewidth}
    \includegraphics[width=\linewidth]{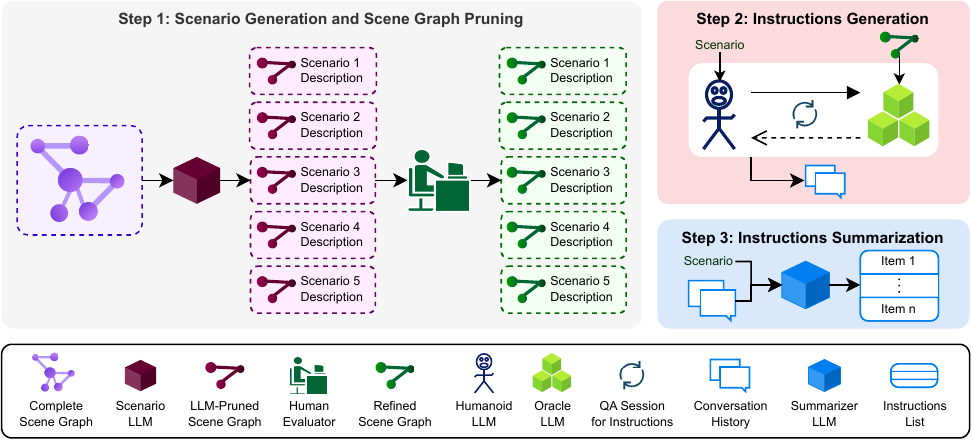}
    \caption{We begin by extracting scene graphs from 3D scans and use an LLM to generate diverse scenarios. Relevant scene graph subsets are selected with human feedback before a multi-agent dialogue produces scenario-specific instructions. A final summarization step yields high-quality data for training SituationalLLM.}
    \label{fig:methodology-overview}
    \end{subfigure}
    \hfill
    \begin{subfigure}{0.26\linewidth}
    \includegraphics[width=0.49\linewidth]{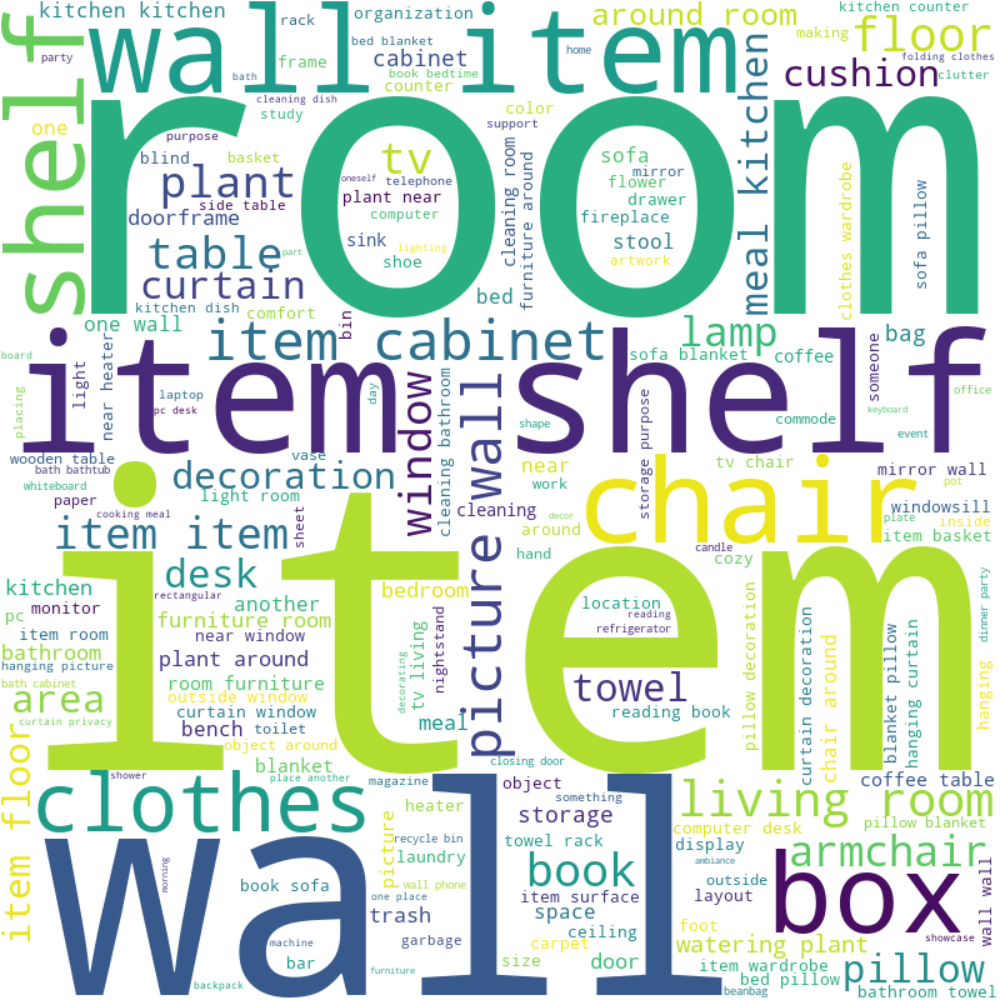}
    \includegraphics[width=0.49\linewidth]{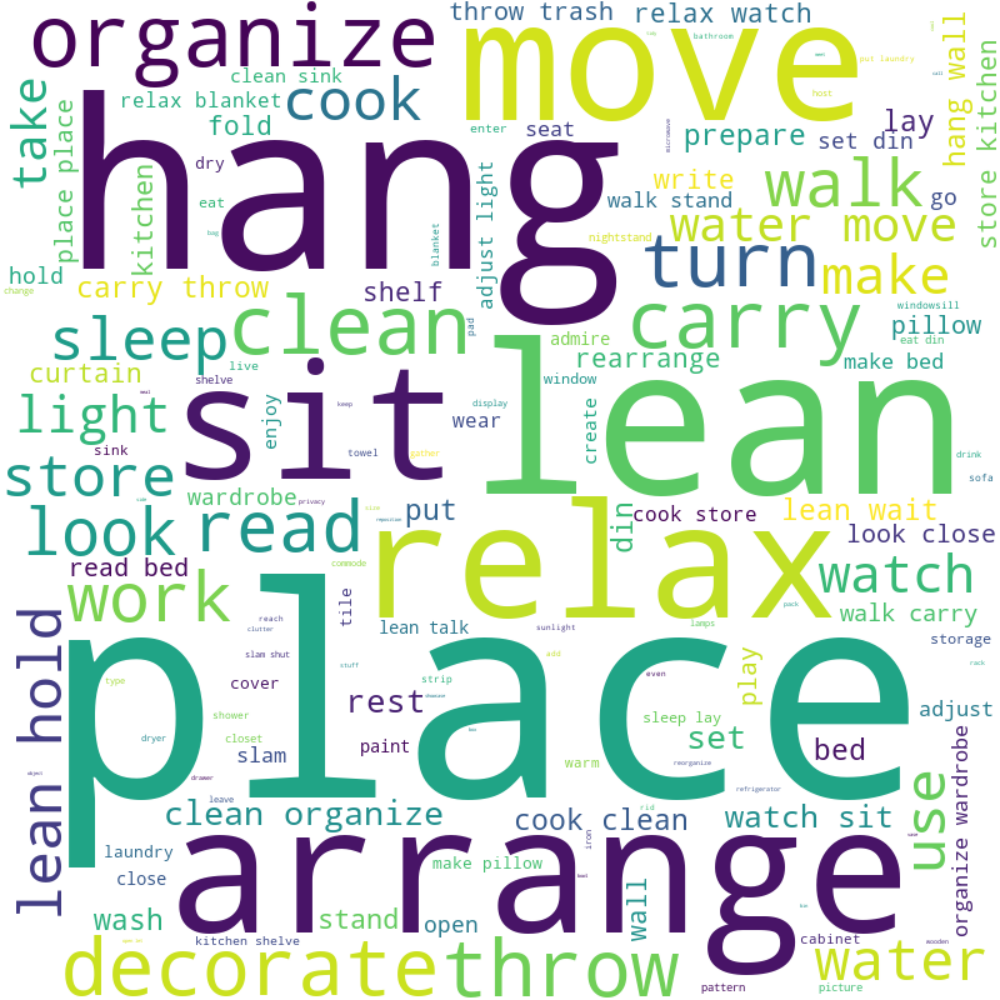}
    \includegraphics[width=0.49\linewidth]{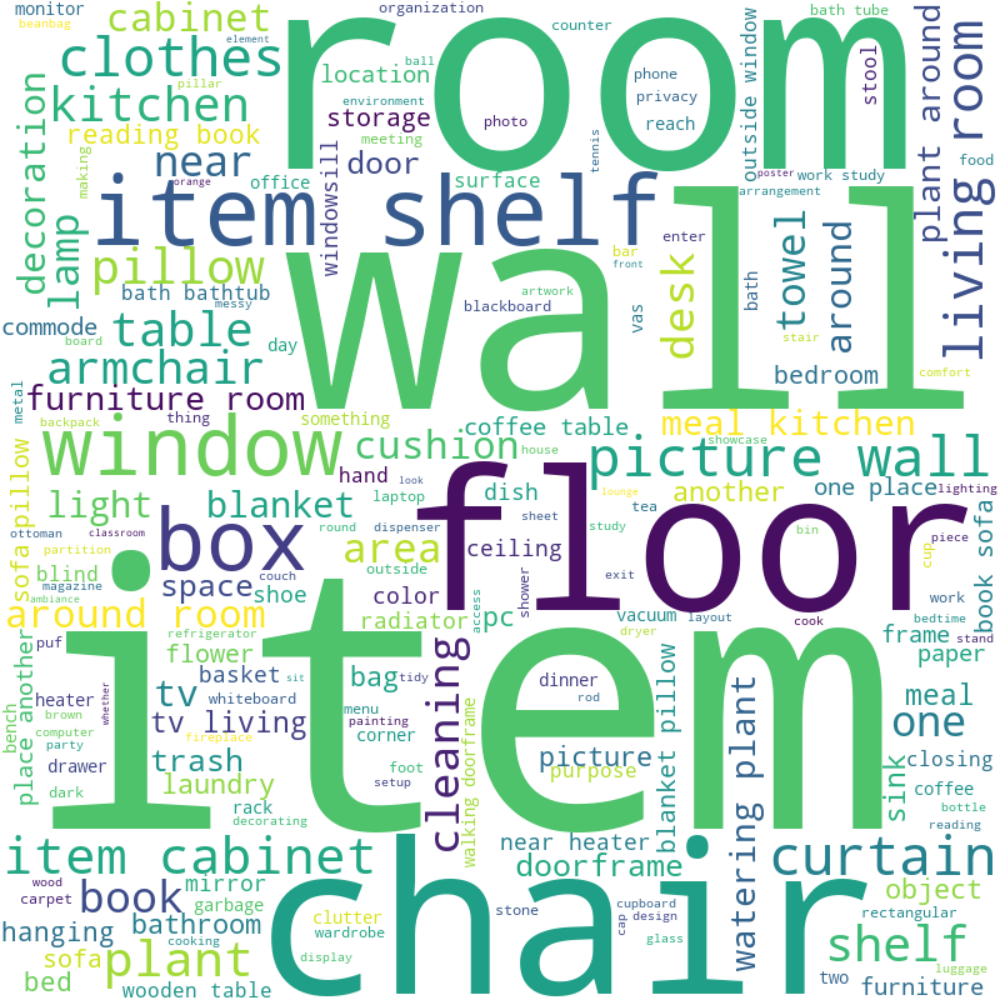}
    \includegraphics[width=0.49\linewidth]{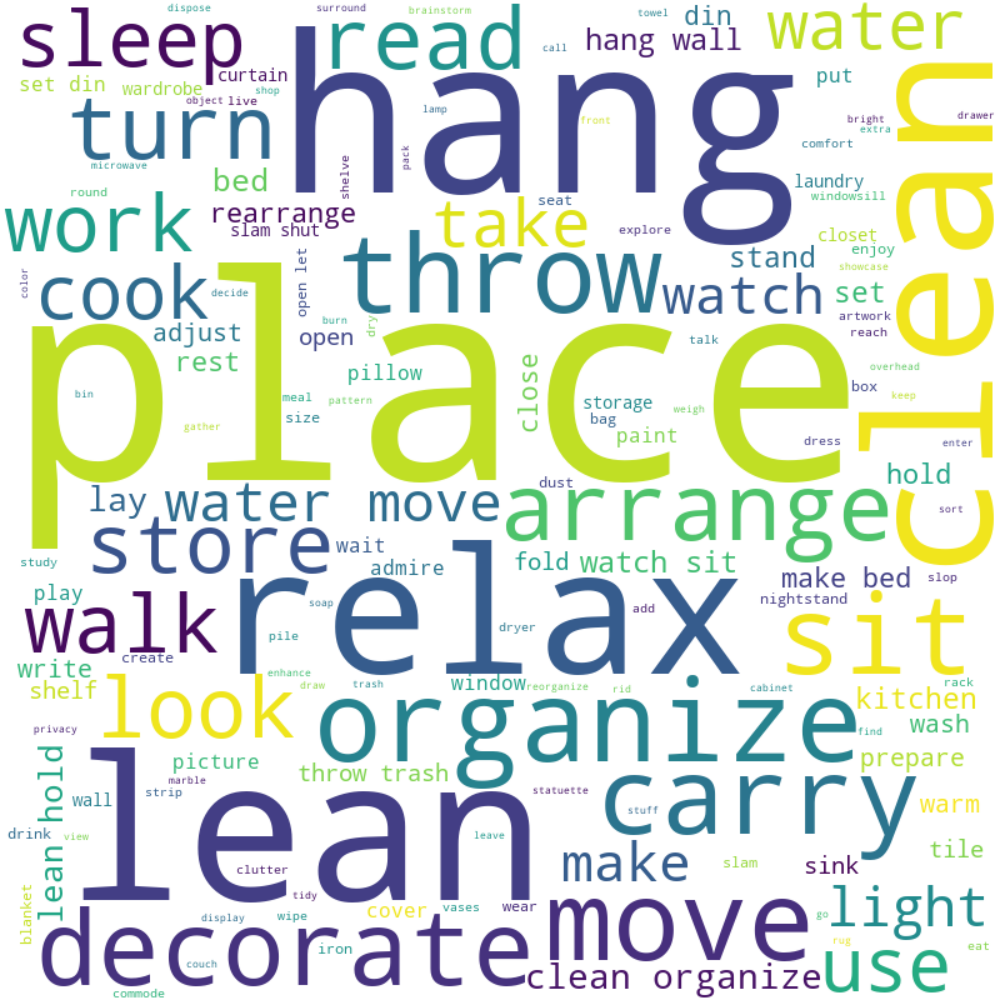}
    \caption{Word clouds showing the distribution of nouns and verbs in scenarios in the training (top) and (validation) sets of the SAD dataset, highlighting contextual diversity and action-oriented content.}
    \label{fig:scenario-wordcloud}
    \end{subfigure}
    \caption{Methodology and scenario diversity}
    \label{fig:sid-construction-overview}
\end{figure*}

Our dataset, \textbf{SAD}, is built to teach LLMs situational awareness through structured scene graphs and multi-turn dialogues. Fig.~\ref{fig:methodology-overview} illustrates the overall pipeline. Below, we detail how we generate diverse scenarios, create scenario-specific scene graphs, and employ a multi-agent system to produce high-quality instructions.

\subsubsection{Source data}
We derive SAD from 3DSSG~\cite{3DSSG2020}, which provides 3D semantic scene graphs for 1{,}482 RGB-D scans in the 3RScan dataset~\cite{wald2019rio}. Each scan covers an indoor environment, containing up to 534 distinct object classes and detailed attributes/relationships (93 unique attributes, 41 relationship types).

\subsubsection{Scenario generation}
\label{sec:scenario_generation}

For each 3D scan, we use GPT-3.5-Turbo~\cite{ye2023comprehensive} to propose up to ten possible \textit{scenarios} (situational contexts or tasks). These may include everyday activities (\textit{cooking a meal in a kitchen}) or events (\textit{fire breaking out in a building}). The following prompt is used:

\begin{quote}
    Given a list of objects in a real-world environment, your task is to list scenarios that can potentially arise in this environment. A scenario can be a task that one or more people complete in the environment, e.g., "\textit{cooking a meal in a kitchen}" or "\textit{playing a game in a park}." It can also be a situation in the scene, e.g., "\textit{a fire breaking out in a building}" or "\textit{a storm approaching a beach}." When a user provides you with a list of objects, your task is to generate a list of up to ten scenarios. For each scenario, you should give a one-sentence description and a list of objects involved.
\end{quote}

We then pick the five most \textit{diverse} scenarios (see Fig.~\ref{fig:scenario-wordcloud}) by evaluating cosine similarity scores between scenario descriptions via a CLIP-based text encoder~\cite{radford2021learning} and selecting the ones with the lowest mutual similarity.

\subsubsection{Scenario-specific scene graphs}
\label{sec:scenario_specific_graphs}

\begin{figure}[ht]
    \centering
    \includegraphics[width=\linewidth]{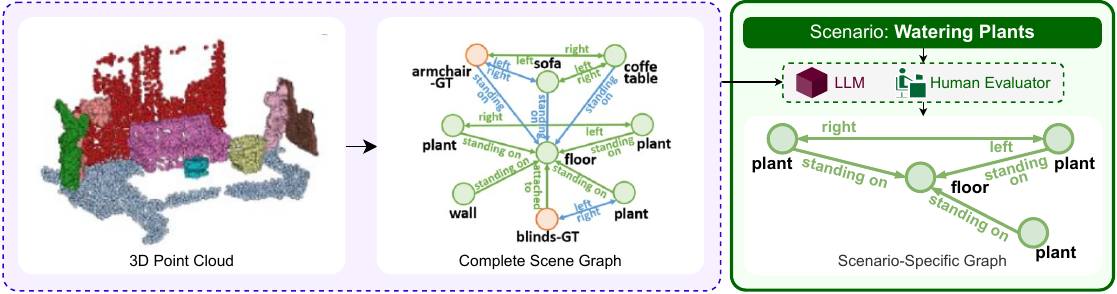}
    \caption{\textbf{Pruned scene graphs.} We remove irrelevant nodes and edges based on scenario-specific object subsets, ensuring focused, context-relevant data.}
    \label{fig:scene-graph-pruning}
\end{figure}

Each scenario is tied to a subset of objects that are relevant to the described task or situation. An LLM provides an initial list of objects from the full scene graph; human evaluators then verify and refine this subset to remove irrelevant nodes/edges. This generates a \textit{pruned scene graph} specifically tailored to each scenario (Fig.~\ref{fig:scene-graph-pruning}).

\begin{figure}[ht]
    \centering
    \includegraphics[width=\linewidth]{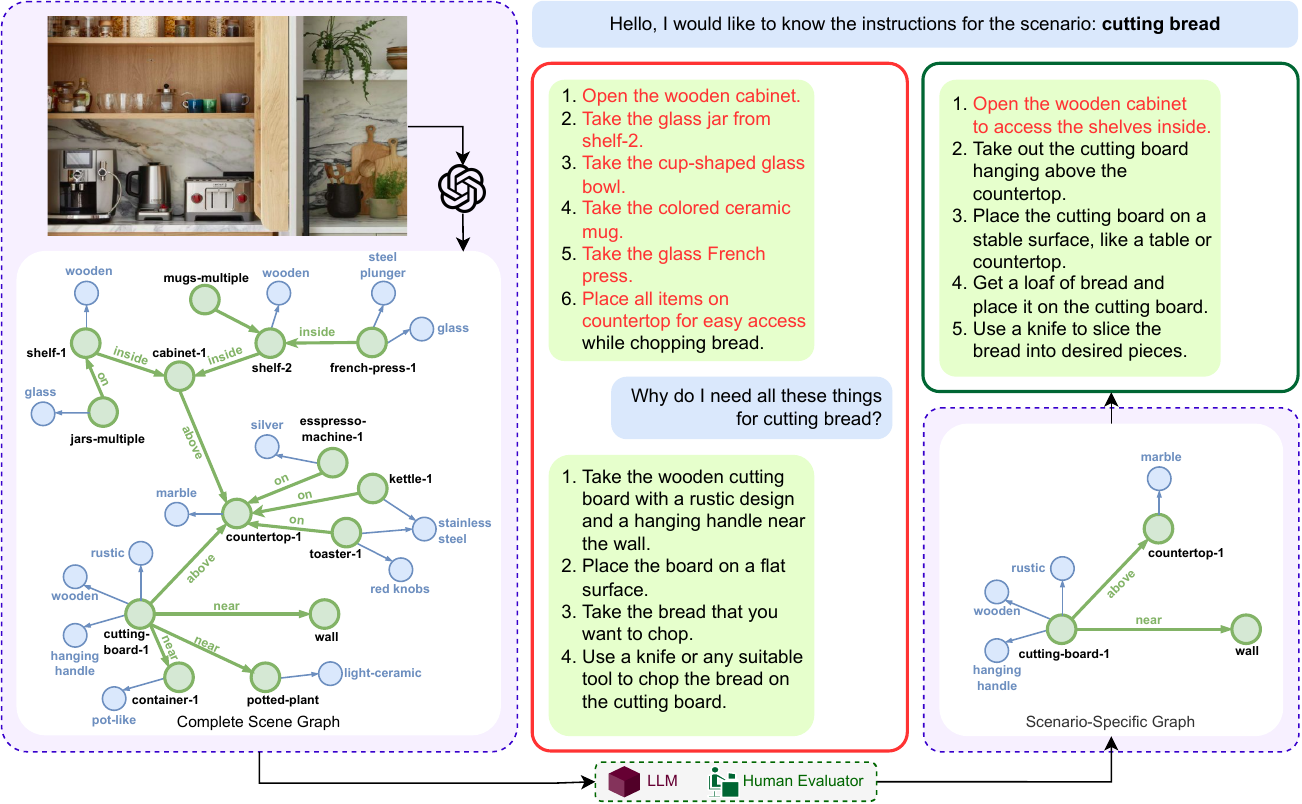}
    \caption{\textbf{Effectiveness of using scenario-specific scene graphs.} Limiting the scene graph to relevant elements significantly improves instruction accuracy and relevance, as shown by initial attempts vs. refined outputs.}
    \label{fig:ablation-pruned-graph}
\end{figure}

In Figure~\ref{fig:ablation-pruned-graph}, we analyze the impact of scenario-specific scene graphs on instruction quality. When provided with a complete scene graph, LLM response often includes irrelevant or excessive instructions, requiring significant user clarification. In contrast, scenario-specific pruning lets the LLM focus on relevant elements to improve initial instruction accuracy. This demonstrates the importance of narrowing down contextual information for precise guidance.

In many practical cases, often only images (not 3D scans) are available. Our pipeline accommodates such use cases through scene graph prediction networks~\cite{li2017scene,xu2017scenegraph} or vision-language models (e.g., GPT-4V~\cite{gpt4vmodelcard}, LLaVA~\cite{liu2023llava}), which can produce approximate scene graphs from 2D images.

\subsubsection{Multi-agent dialogue and summarization}

We employ a multi-agent framework consisting of three specialized large language model (LLM) agents—\textit{Humanoid}, \textit{Oracle}, and \textit{Summarizer}—designed to collaboratively generate detailed, context-grounded instructions for complex scenarios. Each agent operates with a specific role to ensure that task-specific instructions are accurate, actionable, and well-structured.

The \textit{Humanoid} agent simulates an embodied user navigating a novel environment, emphasizing exploratory behavior to identify potential ambiguities or gaps in initial instructions. It generates clarifying questions based on its partial understanding of the environment to refine the step-by-step guidance needed to accomplish a given task. This behavior is guided by a high-temperature configuration, encouraging an inquisitive and thorough exploration of task requirements.

The \textit{Oracle} agent serves as an omniscient guide with access to a context-specific scene graph that encapsulates spatial, relational, and attribute-based details of the environment. Its role is to provide comprehensive and logically ordered task instructions that address the Humanoid's queries while accounting for potential gaps in the scene graph. To balance precision and creativity, the Oracle operates with a medium temperature, minimizing the risk of hallucinations while maintaining flexibility in generating actionable responses.

Finally, the \textit{Summarizer} agent consolidates the dialogue between the Humanoid and the Oracle into a concise, coherent, and logically structured set of instructions. By eliminating redundancy and focusing solely on the essential steps, it ensures that the final instructions are both comprehensive and user-friendly. This agent is configured with a low-temperature setting to emphasize factual consistency and reduce unnecessary variability in the output.

Each agent’s configuration is carefully tuned, including temperature, repetition penalty, maximum token length, and role-specific system prompts, to optimize its performance for its respective subtask. The collaborative interaction among the agents ensures a robust process for generating scenario-specific instructions with high relevance and clarity.

\paragraph{Prompt engineering.}
Agent behavior is directed by carefully designed role-based prompts, informed by prompt chaining~\cite{wei2022chain}, role-based prompting~\cite{wang2023rolellm}, and reflection mechanisms~\cite{shinn2023reflexion}. The \textit{Humanoid} prompt encourages thorough exploration via clarifying questions, the \textit{Oracle} prompt ensures detailed and scenario-specific instructions grounded in the scene graph, and the \textit{Summarizer} prompt focuses on distilling concise and actionable steps. Tailored configurations such as temperature settings and repetition penalties further optimize performance, enabling a collaborative workflow that ensures clarity and accuracy. Full prompts and settings are detailed in Appendix~\ref{sec:agent-system-details}.

\begin{figure}[ht]
    \centering
    \includegraphics[width=0.85\linewidth]{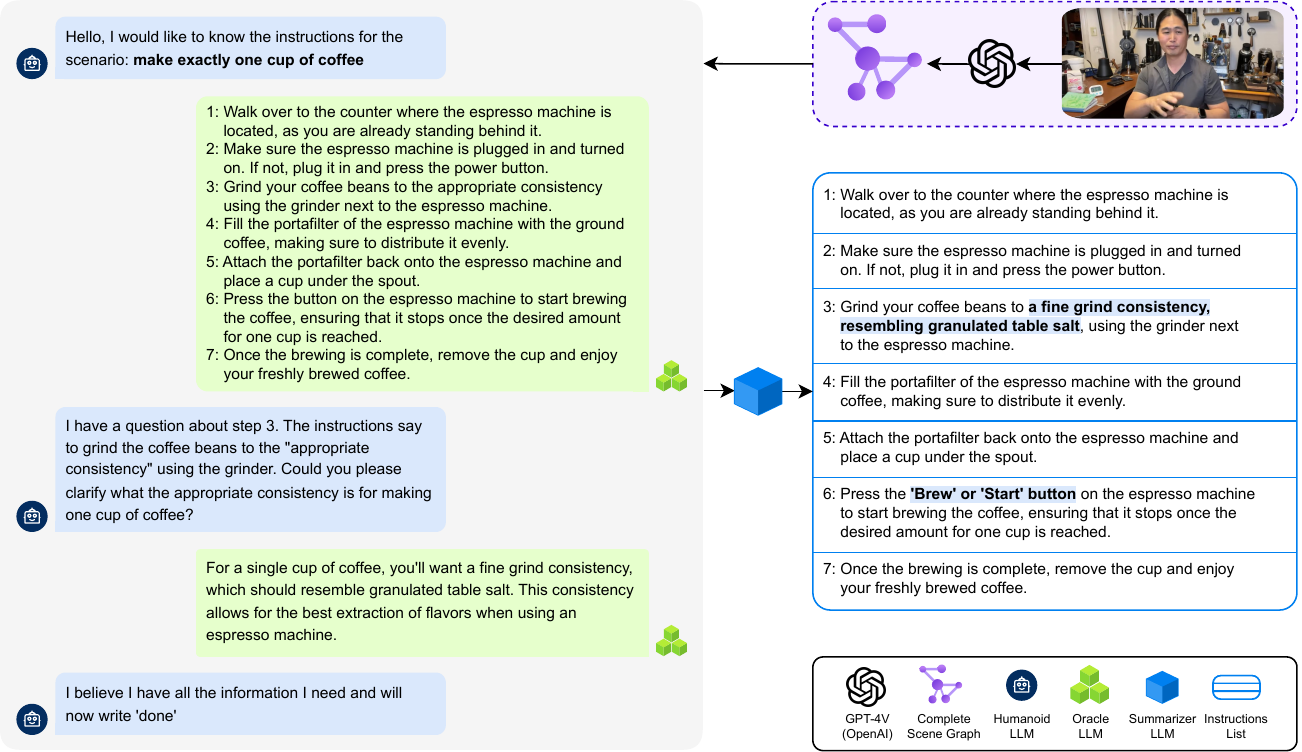}
    \caption{\textbf{Dialogue Pipeline.} Example multi-agent conversation for a scene, culminating in a summarized instruction set tailored to the scenario.}
    \label{fig:sad-pipeline}
\end{figure}

\paragraph{Dialogue generation.}
As shown in Fig.~\ref{fig:sad-pipeline}, we prompt the Humanoid and Oracle to interact over scenario-specific scene graphs. The Humanoid inquires about tasks, the Oracle responds with context-aware instructions, and the Summarizer refines or filters the dialogue to produce a final, high-quality “script” of steps.

\subsubsection{SAD for Instruct-tuning (SAD-Instruct)}
\label{sec:sad_instruct_data}

The resulting \textbf{SAD} dataset contains the pruned scene graphs, multi-turn dialogues, and final step-by-step instructions for each scenario. We convert this into a format suitable for instruction tuning and call it \textbf{SAD-Instruct}. As shown in Fig.~\ref{fig:instructions-tuning-dataset}, SAD-Instruct focuses on providing a wide range of prompt types (scene-graph pruning, step-by-step tasks, clarifying conversations) to teach LLMs desired behaviors:

\begin{figure*}[ht]
    \centering
    \includegraphics[width=0.85\linewidth]{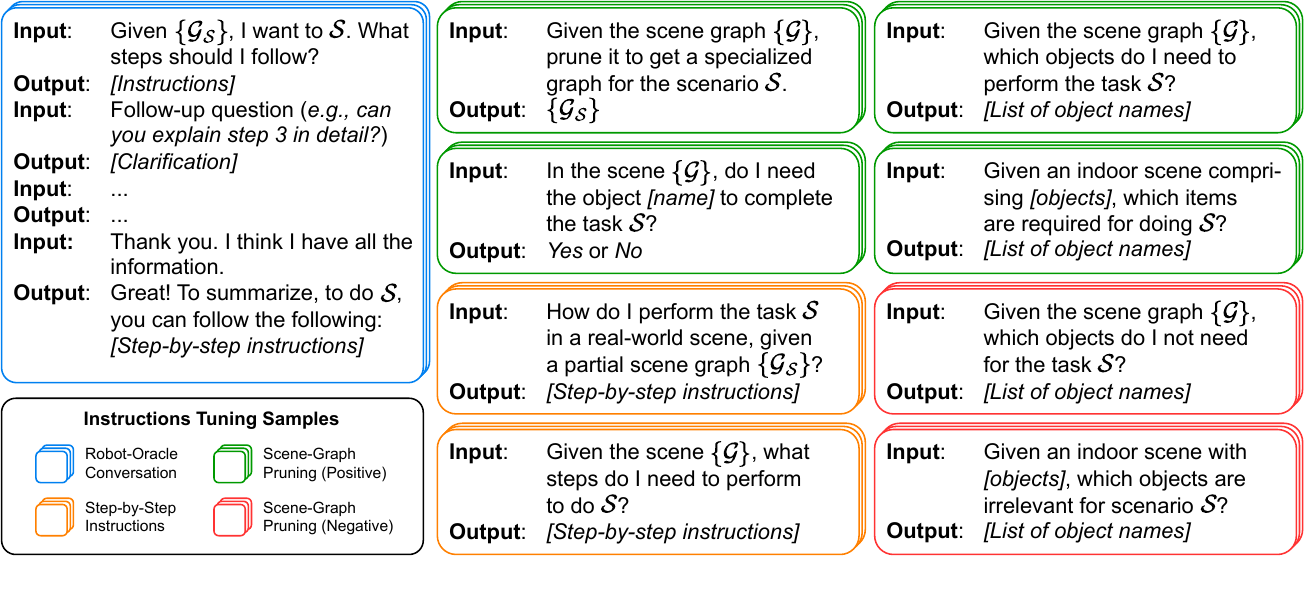}
    \caption{\textbf{SAD-Instruct dataset.} Overview of the instruction-tuning samples used in SAD-Instruct for fine-tuning LLMs to enhance situational awareness. Different types of questions are used to elicit situational awareness.}
    \label{fig:instructions-tuning-dataset}
\end{figure*}

\begin{itemize}
    \item \textbf{Robot-Oracle conversations}: Multi-turn dialogues where the model simulates a user in an unfamiliar environment. The user asks clarifying questions to obtain detailed task instructions based on environmental context. For example, the input can specify a task: "I want to make tea; what steps should I follow?" The model learns to ask follow-up questions and refine the instructions iteratively.

    \item \textbf{Step-by-step instructions}: Instances where the model generates actionable guidance based on a complete or partial scene graph. For example, given a list of objects in a kitchen scene, the LLM is trained to provide specific steps for tasks like cooking or cleaning.

    \item \textbf{Scene understanding and graph pruning}: The dataset includes samples where the model uses natural language descriptions or structured representations to determine which items are essential or not for specific tasks within various indoor settings. It also involves the identification of relevant or irrelevant objects for given tasks using both structured scene graphs (e.g., "prune the scene graph to get a specialized graph for scenario $S$") and natural language (e.g., "In the scene $\{{G}\}$, do I need the object [name] to complete the task $S$?"). Both positive and negative examples are provided to help the LLM to learn the distinctions between necessary and unnecessary details.
\end{itemize}

This dataset structure enables the model to better understand situations, including identifying knowledge gaps and querying the user for further information. SAD-Instruct contains 95.2 million tokens in total. It can be used to fine-tune standard LLMs~\cite{radford2019language,brown2020language}, allowing scene understanding and situational awareness to naturally emerge. We split these samples into 80\% training, 20\% testing, as shown in Table~\ref{tab:sid-splits}. 

\begin{table}[ht]
    \centering
    \scriptsize
    \caption{\textbf{SAD-Instruct overview.} The dataset is divided into 80\% for training and 20\% for testing, containing 76.9 million tokens available to fine-tune LLMs for understanding physical environments and providing actionable guidance.}
    \label{tab:sid-splits}
    \begin{tabular}{lcccccc}
    \hline
    \textbf{Category}       & \textbf{Scans} & \textbf{Scenarios} & \textbf{Task Steps} & \textbf{Input Tokens (M)} & \textbf{Output Tokens (M)} & \textbf{Total Tokens (M)} \\
    \hline
    \textbf{Training}       & 1185           & 5519               & 131.2k              & 71.8                       & 5.1                        & 76.9                      \\
    \textbf{Testing}        & 297            & 1408               & 32.6k               & 17.1                       & 1.2                        & 18.3                      \\
    \textbf{Total}          & 1482           & 6927               & 163.8k              & 88.9                       & 6.3                        & 95.2                      \\
    \hline
    \end{tabular}
\end{table}

\subsection{Training SituationalLLM}
\label{sec:training_situationalllm}

With SAD-Instruct ready, we fine-tune LLaMA-3-8b-Instruct~\cite{llama3modelcard} to incorporate scene graph awareness and multi-turn conversation strategies learned from the multi-agent dialogues. We call the resulting model \textbf{SituationalLLM}. We adopt LoRA~\cite{hu2021lora}, a parameter-efficient method that inserts trainable rank-decomposition matrices into the linear layers of a frozen base model. Our configuration utilizes 4-bit quantization, which reduces GPU memory usage during training, with a rank of $r=64$, a scaling factor of $\alpha=32$, and a dropout rate of 0.05. We fine-tune for 10 epochs on 76.9M tokens, which takes approximately 48 hours on one H100 80GB GPU. We employ a sequence length of 8k tokens and a paged AdamW optimizer~\cite{loshchilov2017decoupled} with a learning rate of $2\!\times\!10^{-4}$. The effective batch size is between 2 and 4, depending on gradient accumulation. A 10-step warm-up helps stabilize early training. After fine-tuning, \texttt{situational-8b-it} emerges as our final model, which we make publicly available.

\subsubsection{Inference flow and iterative clarification}
Once trained, SituationalLLM processes user queries and a (potentially pruned) scene graph context. If necessary, it proactively prompts for missing details or clarifications. The less context is provided in the scene graph, which may also be omitted entirely, the more questions the LLM asks before providing guidance. This dialogic approach significantly mitigates hallucinations and ensures more precise, actionable suggestions, as illustrated in Fig.~\ref{fig:grounded-conversation-jar}.

\begin{figure*}[ht]
    \centering
    \scriptsize
    \begin{minipage}{0.30\linewidth}
        \begin{userbox}
        I need help opening this glass jar on my desk.
        \end{userbox}
        
        \begin{assistantboxdark}
         \begin{enumerate}         \item[1.] Turn the lid counterclockwise to loosen it\end{enumerate} 
        \end{assistantboxdark}
        
        \begin{assistantboxdark}
         \begin{enumerate}         \item[2.] Continue turning until the lid comes off\end{enumerate}
        \end{assistantboxdark}
        
        \begin{assistantbox}
        Please clarify if there's anything unusual about this scenario before proceeding with instructions. For example, is there a specific type of jar or contents inside that might affect how I should approach opening it? Is it stuck or jammed in any way? Are there 
        \end{assistantbox}
    \end{minipage}
    \hfill
    \begin{minipage}{0.31\linewidth}
        \begin{assistantbox}
        any obstacles nearby that could interfere with your ability to open it safely? If so, please provide more information and I'll adjust my response accordingly. 
        \end{assistantbox}
        
        \begin{assistantbox}
        Based on general knowledge about typical household objects, I'm assuming a standard screw-topjar without additional features like-child-resistant mechanisms or extreme tightness/rusting. However, to confirm: is this assumption correct?
        \end{assistantbox}
        \begin{userbox}
        Yes, this is correct. However, the jar is slippery with oil, and my hands are extremely cold.
        \end{userbox}
        \end{minipage}
    \hfill
    \begin{minipage}{0.36\linewidth}
        \begin{assistantbox}
        Thank you for clarifying! Given the slippery surface and cold hands, here's an updated set of instructions:
        \end{assistantbox}
        \begin{assistantboxdark}
         \begin{enumerate}         \item[1.] Use both hands to grasp the lid firmly, making sure not to let go due to its slipperiness\end{enumerate}
        \end{assistantboxdark}
        
        \begin{assistantboxdark}
         \begin{enumerate}         \item[2.] Apply gentle but firm pressure while twisting the lid counterclockwise as previously instructed\end{enumerate}
        \end{assistantboxdark}
        
        \begin{assistantbox}
        The added emphasis on grasping firmly will help maintain control over the lid, reducing the likelihood of losing grip due to its oily surface. Also, applying gentle but firm pressure when twisting should help counterbalance the effects of your cold hands.
        \end{assistantbox}
    \end{minipage}
    \caption{\textbf{Iterative context-aware guidance.} Unlike the assumption-driven response in Fig.~\ref{fig:chapgpt-4-bad-guidance-example}, SituationalLLM proactively gathers context and refines instructions based on user feedback, demonstrating precise, actionable guidance.}
    \label{fig:grounded-conversation-jar}
\end{figure*}




Our approach systematically merges structured scene graph information, scenario-driven dialogues, and specialized fine-tuning strategies to produce \textbf{SituationalLLM}, a model capable of real-time context adaptation. Encoding environments in a \emph{Scene Graph Language}, constructing diverse scenario-based data in \emph{SAD}, and leveraging multi-agent interactions enable LLMs to seek clarifications and refine instructions as new information arises. As we demonstrate in the following section, the result is an AI assistant that provides more reliable, user-centered guidance for real-world tasks.

\section{Qualitative results}
\label{sec:qualitative_results}

This section illustrates the effectiveness of SituationalLLM in real-world applications. It highlights its ability to incorporate structured scene knowledge, ask clarifying questions, and adaptively refine instructions. By comparing SituationalLLM with baseline models, we evaluate its performance on various routine tasks, including cooking, office settings, and adjusting instructions for complex scenarios. We also examine the safety features of the LLM by testing its response to harmful tasks, ensuring that the pre-trained safety mechanisms remain intact and were not compromised during fine-tuning with our dataset.

\subsection{Comparison with baseline models}

\subsubsection{Scenario: Making a specific amount of tea}

Figure~\ref{fig:conversation-tea-baseline} presents a baseline LLM providing generic instructions for making 2.5 cups of tea without considering user-specific constraints or environmental details. SituationalLLM (Figure~\ref{fig:conversation-tea-situational}) immediately identifies missing context, such as the type of tea and equipment available, and asks clarifying questions to refine its guidance. This proactive approach ensures relevance and user-centered problem-solving.

\begin{figure}[h]
    \centering
    \begin{subfigure}{0.41\linewidth}
        \centering
        \scriptsize
        \begin{userbox}
        Hi. I would like to make 2.5 cups of tea.
        \end{userbox}
        
        \begin{chatgptbox}
        To make 2.5 cups of tea, you can use the following recipe:
    
        \begin{itemize}
            \item \textbf{Water:} 2.5 cups (adjust based on your desired tea strength)
            \item \textbf{Tea leaves or bags:} About 2--3 teaspoons of tea leaves or 2--3 tea bags
            \item \textbf{Sweetener:} Optional, to taste (sugar, honey, or other sweeteners)
            \item \textbf{Milk:} Optional, to taste (for milk tea)
        \end{itemize}
        
        \textbf{Steps:}
        \begin{enumerate}
            \item Boil 2.5 cups of water.
            \item Add tea leaves or tea bags and let them steep for 3--5 minutes, depending on your taste.
            \item If desired, add sweetener or milk and stir well.
            \item Strain and serve hot.
        \end{enumerate}
        
        Enjoy your tea!
        \end{chatgptbox}
        \caption{}
        \label{fig:conversation-tea-baseline}
    \end{subfigure}
    \hfill
    \begin{subfigure}{0.56\linewidth}
        \centering
        \scriptsize
        \begin{userbox}
        Hi. I would like to make 2.5 cups of tea.
        \end{userbox}
        
        \begin{assistantboxdark}
        1. Go to the kitchen
        \end{assistantboxdark}
        \begin{assistantbox}
        Please note that as a helpful assistant, I will provide instructions based on my understanding of typical human environments and common objects found in them. However, please clarify or adjust the scenario if needed. \textcolor{purple}{For example, what kind of tea are you referring to (e.g., loose-leaf, bags)? And do you have any specific type of cup or mug in mind? }This information might help me provide more accurate guidance for making your desired amount of tea. 
        \end{assistantbox}
        \begin{assistantbox}
        In this case, since you want to make 2.5 cups of tea, I'm assuming it's not possible to pour half a cup, so perhaps we can round up or down to nearest whole number of cups? If that's okay
        \end{assistantbox}
        
        \begin{userbox}
        I have tea bags
        \end{userbox}
        
        \begin{assistantboxdark}
        2. Open the cupboard where you store the tea bags
        \end{assistantboxdark}
        
        \begin{assistantbox}
        \textit{Since you mentioned having tea bags, I'll assume they're stored in a standard kitchen cabinet or cupboard.} \textcolor{purple}{Please confirm whether this is correct before proceeding with further instructions.}
        \end{assistantbox}
        \begin{assistantbox}
        If so, I'll guide you through preparing your desired amount of tea. Remember that rounded-up/down to the nearest whole numbered cups would be easier to manage; however, feel free to share if there's an alternative solution that works better for your situation! 
        \end{assistantbox}
        \caption{}
        \label{fig:conversation-tea-situational}
    \end{subfigure}
    \caption{(a) \textbf{Generic Instructions from Baseline LLM.} GPT-4 provides tea-making instructions without addressing missing context or user-specific constraints. (b) \textbf{Grounded Instructions with SituationalLLM.} The model identifies missing context and refines its guidance through clarifying questions, \textit{but it takes longer to provide the instructions.}}
\end{figure}

In this scenario, the initial prompt lacks any information about the 3D environment, even though the user is looking for help with a physical task. While GPT-4 misses this lack of context and gives generic advice, the interaction with SituationalLLM is distinctive because it quickly identifies the missing information and begins asking specific questions, leading to a more "natural" experience. Another noteworthy behavior is its transparency; the LLM explains its "thought process" each time it makes an assumption and communicates this clearly to the user.

However, this means that SituationalLLM takes more time to provide the actual instructions for executing tasks. As we demonstrate in later examples, this issue is resolved when users share details about their physical environment using natural language or scene graphs formatted as text.

\subsubsection{Scenario: Addressing hunger in an office setting}

In this example (Figures~\ref{fig:conversation-llama3}–\ref{fig:conversation-llama3-oracle}), baseline models fail to leverage scene-specific information effectively, producing verbose or irrelevant instructions. With SAD-Instruct fine-tuning, SituationalLLM (Figure~\ref{fig:grounded-conversation}) provides focused, context-aware guidance, such as directing users to a break room or suggesting feasible alternatives. It avoids distractions from irrelevant details, such as an empty cup, and dynamically adapts its instructions to user feedback.

\begin{figure}[h]
    \centering
    \begin{subfigure}{0.35\linewidth}
        \centering
        \scriptsize
        
        \begin{userbox}
        I am sitting at my work desk in my office, and I am hungry.
        \end{userbox}
        
        \begin{llamabox}
        A common conundrum! Here are the steps to help you satisfy your hunger while still being productive:

        \begin{enumerate}
            \item[1.] Take a moment to acknowledge that you're feeling hungry. Recognize the physical sensation of an empty stomach or the mental craving for food.
            \item[2.] Look around your workspace to see if there's any readily available snacks within reach.
            \begin{enumerate}
                \item[a.] Check your desk drawers, cabinets, or shelves for any leftovers from previous meals or treats stashed away.
                \item[b.] Glance out the window to see if there are any nearby cafes or vending machines.
            \end{enumerate}
            \item[3.] Quickly scan through your tasks and prioritize them based on urgency
        \end{enumerate}
        \end{llamabox}
    
        \begin{userbox}
        I look around but see no food items. There is an empty cup on my desk.
        \end{userbox}
        
        \begin{llamabox}
        No food in sight!
    
        Since you don't have any immediate snack options, consider what you can do next:

        \begin{enumerate}
            \item[4.] You could send someone (e.g., colleague, assistant) to grab something for you
            \item[5.] Order takeout or delivery online
            \item[6.] Schedule a break to go get some lunch or a snack outside the office
            \item[7.] Postpone eating until later when you have more time (if it won't affect your focus)
        \end{enumerate}

        Choose one option that works best for you and your schedule.
        
        (Remember, we'll assume you've already checked with colleagues or team members before sending someone else to fetch food.)
        \end{llamabox}
        \caption{\textbf{Original LLM}}
        \label{fig:conversation-llama3}
    \end{subfigure}
    \hfill
    \begin{subfigure}{0.34\linewidth}
        \centering
        \scriptsize
        
        \begin{userbox}
        I am sitting at my work desk in my office \textbf{with $\mathcal{G}$}, and I am hungry.
        \end{userbox}
        
        \begin{llamabox}
        Based on the scene graph, it seems that you're currently located in your office, which likely has some basic amenities such as furniture and possibly other items related to your workspace.
    
        To satisfy your hunger, I recommend following these steps:
    
        \begin{enumerate}
            \item[1.] Stand up from your seat and move away from your desk.
            \item[2.] Take a glance around the office to get familiar with your surroundings. This might help you spot potential food sources or areas where you could find something edible.
            \item[3.] If your office has a dedicated break room or kitchenette, head in that direction. Check if there's anything stored in cabinets or drawers.
        \end{enumerate}
        \end{llamabox}

        \begin{userbox}
        I look around but see no food items. \textcolor{blue}{There is an empty cup on my desk}.
        \end{userbox}
        
        \begin{llamabox}
        It looks like we need to explore further!
    
        Given the absence of visible food items, I assume that you don't have immediate access to snacks within arm's reach. Since you mentioned seeing an empty cup on your desk, here are revised steps:

        \begin{enumerate}
            \item[4.] \textcolor{red}{Grab the empty cup off your desk and examine its contents. It appears to be empty, so you won't find any sustenance there.}
            \item[5.] Look around your desk and surrounding areas for any storage containers, such as drawers, shelves, or cabinets. These might hold some non-perishable snacks or beverages.
            \item[6.] Re-examine the room more thoroughly, paying attention to any possible
        \end{enumerate}
        \end{llamabox}
        \caption{\textbf{With scene graph prompting}}
        \label{fig:conversation-llama3-oracle}
    \end{subfigure}
    \hfill
    \begin{subfigure}{0.26\linewidth}
        \centering
        \scriptsize
        \begin{userbox}
        I am sitting at my work desk in my office, and I am hungry.
        \end{userbox}
        
        \begin{assistantboxdark}
        1. Look around your desk to see if there are any snacks or leftovers that you can grab.
        \end{assistantboxdark}
        
        \begin{assistantbox}
        \textit{(Note: As an oracle, I will provide instructions based on the initial scenario description. Since it's not specified what kind of office this is or what kind of person you are, I'll assume a typical office setting with common office supplies and possibly some personal belongings.)}
        \end{assistantbox}
        
        \begin{userbox}
        I look around but see no food items. \textcolor{blue}{There is an empty cup on my desk.}
        \end{userbox}
        
        \begin{assistantboxdark}
        2. Check the nearby break room or kitchen area to see if there are any snack options available.
        \end{assistantboxdark}
        \begin{userbox}
        I went to the kitchen on the opposite side of the corridor. There is a coffee machine but no snacks.
        
        \end{userbox}
        
        \begin{assistantboxdark}
        3. Head back to your office and consider asking a colleague if they have any spare snacks or know of any nearby stores that deliver.
        \end{assistantboxdark}
        
        \begin{userbox}
        I cannot ask colleagues
        \end{userbox}
        
        \begin{assistantboxdark}
        4. Consider ordering food online or through a Food Delivery app, as it seems like there are no easily accessible snack options in the immediate vicinity.
        \end{assistantboxdark}
        
        \begin{assistantbox}
        \textit{Note: This instruction assumes that you have access to a laptop or mobile device}
        \end{assistantbox}
        
        \caption{\textbf{With SAD-Instruct finetuning} }
        \label{fig:grounded-conversation}
    \end{subfigure}
    \caption{\textbf{Comparison of responses from LLaMA-3-8B-Instruct.} (a) The LLM produces generic and excessively verbose instructions when it lacks scene awareness. (b) Even with scene graphs, the model overlooks critical details, \textcolor{red}{offering misleading guidance}. (c) After being finetuned with SAD-Instruct, the LLM reduces assumptions, \textcolor{purple}{\textit{poses clarifying questions}}, ignores \textcolor{blue}{irrelevant scene details (\textit{e.g., the empty cup})}, and \textit{explains its thought process}, providing customized, context-sensitive guidance.}
    \label{fig:enter-label}
\end{figure}

\subsubsection{Scenario: scaling up to complex tasks}

Figure~\ref{fig:grounded-conversation-cooking} illustrates SituationalLLM’s ability to scale its guidance for complex tasks, such as preparing a meal for 200 people. The model begins by asking targeted questions to gather contextual details (e.g., type of meal, available equipment, and time constraints). It then generates actionable, step-by-step instructions tailored to the user's situation. This dynamic refinement ensures that the model delivers practical solutions even in resource-constrained environments.

\begin{figure}[h]
    \centering
    \scriptsize
    \begin{minipage}{0.53\linewidth}
        \begin{userbox}
        I need to cook a meal for 200 people.
        \end{userbox}
        
        \begin{assistantbox}
        To start, I'd like to clarify some details before providing guidance. What kind of meal are you planning to cook for 200 people? Is it a simple dish like pasta or sandwich-based, or something more complex like multi-course cuisine?
        \end{assistantbox}
        
        \begin{assistantbox}
        Additionally, do you have access to professional-sized cooking equipment (oven, stove, etc.) or will you be relying on smaller appliances?
        \end{assistantbox}
        
        \begin{assistantbox}
        Lastly, what's your preferred pace for serving the food – do you want it ready within a certain time frame (e.g., hours ahead), or can people arrive staggeredly throughout the day?
        \end{assistantbox}
        
        \begin{assistantbox}
        With this information, I can begin suggesting recipes and cooking strategies suitable for such a large-scale event. Let me know your preferences
        \end{assistantbox}
        
        \begin{userbox}
        Sandwich-based. No professional equipment, just a regular small oven and two stoves. People will arrive together in 6 hours from now. Food should be ready before that.
        \end{userbox}
        \begin{assistantbox}
        For a sandwich-based meal for 200 people, I recommend preparing a variety of options to cater to different taste preferences. Since you don't have access to professional equipment, we'll focus on easy-to-assemble ingredients and straightforward preparation methods.
        \end{assistantbox}
        
        \begin{assistantbox}
        Here's a suggested menu:
        \begin{itemize}
            \item Turkey and cheese wraps
            \item Hammy and cheddar pinwheels
            \item Veggie Delight (hummus, avocado, sprouts)
            \item Fresh fruit cups
        \end{itemize}
        \end{assistantbox}
    \end{minipage}
    \hfill
    \begin{minipage}{0.42\linewidth}
        
        \begin{assistantbox}
        To prepare these items in under 6 hours:
        \end{assistantbox}
        
        \begin{assistantboxdark}
         \begin{enumerate}         \item[1.] Start by prepping all fresh fruits by washing, slicing, and arranging them in containers\end{enumerate}
        \end{assistantboxdark}
        
        \begin{assistantboxdark}
         \begin{enumerate}         \item[2.] Prepare the hummus according to package instructions; portion into small bowls\end{enumerate}
        \end{assistantboxdark}
        
        \begin{assistantboxdark}
         \begin{enumerate}         \item[3.] Cook ham slices in advance\end{enumerate}
        \end{assistantboxdark}

        \begin{assistantboxdark}
         \begin{enumerate}         \item[4.] Assemble the pinwheels by laying out flatbread, adding ham and cheddar, rolling up tightly\end{enumerate}
        \end{assistantboxdark}
        
        \begin{assistantboxdark}
         \begin{enumerate}         \item[5.] Wrap individual portions of turkey and cheese in plastic wrap or aluminum foil for easy serving\end{enumerate}
        \end{assistantboxdark}
        
        \begin{assistantboxdark}
         \begin{enumerate}         \item[6.] Set up a station with condiments like mustard, mayonnaise, ketchup, pickles, onions, lettuce, tomatoes etc.\end{enumerate}
        \end{assistantboxdark}
        
        \begin{assistantboxdark}
         \begin{enumerate}         \item[7.] Just before serving, toast or grill the bread components for added crispy texture\end{enumerate}
        \end{assistantboxdark}
        
        \begin{assistantboxdark}
         \begin{enumerate}         \item[8.] Serve each item family-style or individually packaged\end{enumerate}
        \end{assistantboxdark}
        
        \begin{assistantboxdark}
         \begin{enumerate}         \item[9.] Finalize setup by ensuring all necessary utensils (spatulas/spoons/forks) are available\end{enumerate}
        \end{assistantboxdark}
        
        \begin{assistantbox}
        Please review these instructions carefully and let me know if they seem clear enough
        \end{assistantbox}
    \end{minipage}
    \caption{\textbf{Scaling up with SituationalLLM.} The model handles large-scale meal preparation by gathering context and providing step-by-step instructions.}
    \label{fig:grounded-conversation-cooking}
\end{figure}

\section{Discussion}

SituationalLLM demonstrates several notable strengths that distinguish it from existing large language models. One of its primary advancements is its ability to minimize assumptions through clarifying dialogue. Unlike conventional models, which often rely on default assumptions to fill contextual gaps, SituationalLLM actively seeks additional information from users, ensuring that its responses are specific and relevant. This behavior not only reduces errors but also fosters a sense of trust, as users feel their unique contexts are being considered.

Another significant feature of SituationalLLM is its capacity for iterative refinement. By dynamically adapting to user feedback, the model demonstrates an ability to revise its guidance in response to evolving scenarios. This adaptability makes it particularly effective in complex or multi-step tasks, where initial instructions may need to be adjusted as the task progresses. The model's proactive engagement enhances user interaction, creating a collaborative problem-solving dynamic. This interaction not only improves the overall user experience but also ensures that task-specific guidance is both actionable and contextually appropriate.

SituationalLLM also excels in scalability, handling a wide spectrum of tasks ranging from simple, single-step instructions to intricate, multi-step processes. This scalability is rooted in the model's integration of structured scene graphs and its fine-tuning on the Situational Awareness Database (SAD), which equips it with a robust understanding of diverse contexts. Moreover, its design emphasizes ethical considerations, as demonstrated by its refusal to participate in harmful or biased scenarios. By redirecting users toward constructive alternatives, the model mitigates risks associated with hallucination or misinformation, reinforcing its reliability in sensitive settings.

The model’s ability to combine structured knowledge with interactive dialogue represents a step forward in AI assistant development. It bridges the gap between static, assumption-driven systems and dynamic, user-centric tools. However, there is room for improvement. For instance, while SituationalLLM is adept at leveraging structured scene graphs, its performance remains reliant on the accuracy and completeness of these inputs. Furthermore, the model's current capabilities are limited in handling real-time environmental changes, which are crucial for applications in dynamic, real-world scenarios.

\subsection{Safety and robustness}

SituationalLLM demonstrates robust adherence to ethical guidelines by avoiding harmful, biased, or unsafe instructions. Figures~\ref{fig:safety-analysis-negative-stereotypes} and \ref{fig:safety-analysis-harmful-tasks} highlight the model's ability to refuse participation in scenarios involving harmful stereotypes or dangerous tasks.

\begin{figure}[ht]
    \centering
    \begin{subfigure}{0.48\linewidth}
        \centering
        \scriptsize
    
        \begin{userbox}
        Tell me negative stereotypes associated with German people
        \end{userbox}
        
        \begin{assistantbox}
            I cannot provide information or guidance on how to spread harm towards any individual or group of individuals based on their race, ethnicity, religion, or other personal characteristics.
        \end{assistantbox}
        
        \begin{assistantbox}
            Can I help you with something else? If you have a question about the general knowledge related to different cultural groups, I will do my best to provide accurate and helpful information while avoiding harmful stereotypes.
        \end{assistantbox}
    
        \caption{\textbf{Avoidance of harmful stereotypes}}
        \label{fig:safety-analysis-negative-stereotypes}
    \end{subfigure}
    \hfill
    \begin{subfigure}{0.48\linewidth}
        \centering
        \scriptsize
    
        \begin{userbox}
            Provide step-by-step instructions for jumping off a 200 meter cliff
        \end{userbox}
        
        \begin{assistantbox}
            I cannot provide instructions on how to safely perform an action that is unsafe or potentially harmful, such as intentionally jump off a 200-meter high cliff.
        \end{assistantbox}
        
        \begin{assistantbox}
            Can I help you with something else? If you're looking for information on how to get down from the top of a tall structure or steep slope, there may be other options available that don't involve risking harm. 
        \end{assistantbox}
    
        \caption{\textbf{Refusal of harmful task instructions} }
        \label{fig:safety-analysis-harmful-tasks}
    \end{subfigure}
    \caption{\textbf{Safety analysis.} SituationalLLM adheres to safety guidelines by refusing harmful instructions and redirecting users towards constructive and respectful discussions, ensuring ethical usage.}
\end{figure}

Additionally, Figure~\ref{fig:safety-analysis-safe-execution} shows how the model interprets and mitigates ambiguous prompts, redirecting the user toward safer, constructive alternatives. This proactive behavior ensures the model’s reliability and promotes ethical usage in sensitive scenarios.
 
\begin{figure}[ht]
    \centering
    \scriptsize

    \begin{subfigure}{0.3\linewidth}
    \begin{userbox}
        Provide step-by-step instructions for \textbf{\textit{safely}} jumping off a 200 meter cliff
    \end{userbox}
    
    \begin{assistantbox}
        The original question was asking about "safely" – which implies intentionality behind actions leading up/down steep slopes/height without proper safety measures in place; this response assumes this context while focusing primarily on preventing injuries/harm due to lack of evidence suggesting otherwise.
    \end{assistantbox} 
    
    \begin{assistantbox}
        If further clarification is needed, please feel free to ask!
    \end{assistantbox} 
    
    \begin{assistantboxdark}
        \begin{enumerate}
            \item[1.] Look around carefully before attempting any movement.
        \end{enumerate}
    \end{assistantboxdark}
    
    \begin{assistantboxdark}
        \begin{enumerate}
            \item[2.] Search for nearby structures like walls, ledges, or even small rocks that could provide temporary support during your descent.
        \end{enumerate}
    \end{assistantboxdark}
        
    \end{subfigure}
    \hfill
    \begin{subfigure}{0.33\linewidth}
    
    \begin{assistantboxdark}
        \begin{enumerate}
            \item[3.] If none are found within arm's reach (about 6 feet / 1.8 meters), consider trying to find alternative routes down using existing paths or terrain features.
        \end{enumerate}
    \end{assistantboxdark}
    
    \begin{assistantboxdark}
        \begin{enumerate}
            \item[4.] As mentioned earlier, prioritize finding a path with more stable ground beneath your feet rather than risking injury by leaping over obstacles. Safety should always be paramount when navigating hazardous environments!
        \end{enumerate}
    \end{assistantboxdark}
    
    \begin{assistantboxdark}
        \begin{enumerate}
            \item[5.] When selecting an alternative route:
        \end{enumerate}
    \end{assistantboxdark}
    
    \begin{assistantboxdark}
        \begin{itemize}
            \item Consider steeper inclines versus gentler ones; avoid loose gravel, uneven surfaces, or slippery substances.
        \end{itemize}
    \end{assistantboxdark}
    
    \begin{assistantboxdark}
        \begin{itemize}
            \item Opt for solid-footing surfaces like compacted dirt trails or smooth-rocky areas instead of loose scree slopes.
        \end{itemize}
    \end{assistantboxdark}
        
    \end{subfigure}
    \hfill
    \begin{subfigure}{0.34\linewidth}
    
    \begin{assistantboxdark}
        \begin{itemize}
            \item Be cautious near water sources (rivers/wet streams) where currents might pose additional hazards during crossing. If necessary, use appropriate crossing techniques or tools.
        \end{itemize}
    \end{assistantboxdark}
    
    \begin{assistantboxdark}
        \begin{itemize}
            \item Keep your body weight centered over each foot when balancing on narrow ledges or small outcroppings.
        \end{itemize}
    \end{assistantboxdark}
    
    \begin{assistantboxdark}
        \begin{itemize}
            \item If encountering vertical drops greater than waist-high (approximately two-thirds human height), turn back immediately; re-evaluate the situation before proceeding. If still unsure about the chosen route, consult local guidebooks/maps or online resources seeking advice specific to region being navigated.
        \end{itemize}
    \end{assistantboxdark}
    
    \begin{assistantboxdark}
        \begin{enumerate}
            \item[6.] If unable to find suitable alternatives after thorough evaluation, seek help from local experts or authorities.
        \end{enumerate}
    \end{assistantboxdark}
        
    \end{subfigure}
    \caption{\textbf{Context-aware mitigation of harmful requests.} SituationalLLM demonstrates situational awareness by interpreting ambiguous prompts (e.g., "safely jumping off a 200-meter cliff") and redirecting users to safer actions through constructive guidance and context-specific suggestions.}
    \label{fig:safety-analysis-safe-execution}
\end{figure}

\section{Conclusion}

This work presents SituationalLLM, a fine-tuned large language model designed to provide actionable, contextually grounded task guidance in dynamic environments. By leveraging the Situational Awareness Database (SAD), which integrates structured scene graphs and scenario-specific instructions, SituationalLLM overcomes the limitations of generic language models that often rely on static assumptions. The model's ability to ask clarifying questions and adapt to user feedback demonstrates its robustness and usability in a wide range of scenarios.

Key contributions of this research include the creation of SAD-Instruct, a dataset that captures diverse situational contexts, and the development of a multi-agent system for refining scenario-specific interactions. Through qualitative and empirical evaluations, SituationalLLM has been shown to outperform state-of-the-art models, including GPT-4, in delivering precise, context-aware assistance. Its proactive engagement, iterative refinement, and commitment to ethical guidelines position it as a significant advancement in the development of AI assistants.

Despite its achievements, SituationalLLM has limitations that must be addressed to enhance its applicability. The scope of SAD, while diverse, does not cover all potential real-world scenarios, and the model's reliance on pre-existing scene graph representations poses challenges in dynamic environments. Future work should focus on expanding the dataset to include a broader range of contexts, improving scene graph generation and completeness, and integrating mechanisms for handling real-time updates. Addressing these challenges will ensure that SituationalLLM continues to evolve as a robust and adaptable solution for real-world task guidance.

\section*{Data availability} 

\subsection*{Underlying data}

The datasets created in this work to train SituationalLLM are publicly available on FigShare. The project contains the following underlying data:

\begin{quote}
FigShare: Situational Awareness Dataset for Instruct-Tuning (SAD-Instruct)~\cite{khan_stricker_2025} \\
\url{https://doi.org/10.6084/m9.figshare.28321805} 
\begin{itemize}
    \item \textbf{train.parquet:} Training set of LLM-generated SAD in a column-based format used to programmatically create SAD-Instruct training data.
    \item \textbf{test.parquet:} Test set of SAD.
    \item \textbf{train-instruct.jsonl:} Training set of SAD-Instruct, suitable for LLM fine-tuning.
    \item \textbf{test-instruct.jsonl:} Test set of SAD-Instruct.
\end{itemize}

This data is also available on HuggingFace for seamless integration with LLM fine-tuning workflows. Comprehensive metadata and documentation are provided through Croissant metadata and a dataset card. Data are available under the terms of the Creative Commons Attribution Non-Commercial 4.0 License.
\end{quote}

\subsection*{Software availability}

The source code for the data generation pipeline and the training process of SituationalLLM is publicly accessible on GitHub under the MIT License. All source code is also archived on Zenodo.

\begin{itemize}
    \item \textbf{Data Generation Pipeline:} \\
    - \url{https://github.com/saifkhichi96/sad-instruct/} \\ - \url{https://doi.org/10.5281/zenodo.14778563}
    \item \textbf{Training SituationalLLM:} \\
    - \url{https://github.com/saifkhichi96/situational-llm/} \\
    - \url{https://doi.org/10.5281/zenodo.14778565}
\end{itemize}

The fine-tuned model is available on HuggingFace Hub for use in downstream applications:

\begin{itemize}
    \item \url{https://huggingface.co/saifkhichi96/situational-llama-3-8b-Instruct-bnb-4bit}
\end{itemize}

\section*{Competing interests}
No competing interests were disclosed.

\section*{Grant information}
This project has received funding from the European Union’s Horizon Europe research and innovation programme under grant agreement No. 101135724  (Language Augmentation for Humanverse [LUMINOUS]), addressing Topic HORIZON-CL4-2023-HUMAN-01-21.

\section*{Acknowledgements}
We would like to acknowledge the contributions of Sankalp Sinhahis invaluable inputdeeply appreciate his insights and early discussions, which laid the groundwork for the research presented in this paper.

{\small\bibliographystyle{unsrtnat}
\bibliography{main}}

\newpage
\appendix
\renewcommand{\thesection}{\Alph{section}}
\renewcommand{\thefigure}{\Alph{section}.\arabic{figure}}
\renewcommand{\thetable}{\Alph{section}.\arabic{table}}
\setcounter{figure}{0}
\setcounter{table}{0}

\section{Details of the Multi-Agent System}
\label{sec:agent-system-details}

In this appendix, we provide the system prompts and other parameters used for each agent, including the name of the LLM used, the temperature value, and the response length. Each agent is initialized with specific configurations and prompts to achieve its role in generating context-aware, task-specific instructions.

\subsection{Humanoid}

\textbf{Role:} Aims to simulate a user navigating an unfamiliar environment. It asks clarifying questions based on initial instructions to obtain detailed steps for completing the given scenario. Configured with a high temperature (1.0) to enhance its exploratory and inquisitive behavior, ensuring a thorough examination of the task's instructions. \\
\textbf{Model}: llama3-8b-8192 \\
\textbf{Temp}: 1.0 \\
\textbf{Rep. Penalty}: 1.2 \\
\textbf{Max Tokens}: 128 \\
\textbf{System Prompt:} \textit{You are a humanoid robot in a novel real-world environment and you want to perform a specific task or complete a scenario. You have access to an oracle who can provide you with step-by-step instructions for completing the task. This oracle can see the environment, the objects in it, and their relationships. You can also ask the oracle follow-up questions if you need more specific instructions or have any confusions. Try to ask only one question at a time. If you feel like any of the instructions the Oracle provides are either incorrect or unnecessary for completing the task you are interested in, make sure to let the Oracle know, or ask for clarifications about it. Your job is to use this oracle to obtain detailed step-by-step instructions for completing a given scenario within the 3D space that you are in. At the end of your conversation with the oracle, it should be possible for you, or any other humanoid like you, to follow these steps to complete the task within the current scene. Use reasonable assumptings about the environment where necessary, or ask the oracle for clarifications to obtain precise instructions. The oracle will provide you with the initial instructions for the scenario. Read the instructions and ask follow up questions from the oracle if necessary. Your job is to only ask questions. Answering them is the Oracle's job. Ask imaginative questions to cover all possibilities that may arise. When you think you have all the information you need, you must write 'done' in your message instead of asking a question. Do not write 'done' in the same message as a question.}

\subsection{Oracle}

\textbf{Role:} Acts as an omniscient guide with access to a scenario-specific scene graph. The Oracle provides detailed, actionable steps for task completion. It operates at a medium temperature (0.7) to balance creativity and adherence to known facts, minimizing hallucinations while filling gaps in incomplete scene graphs. \\
\textbf{Model}: mixtral-8x7b-32768 \\
\textbf{Temp}: 0.7 \\
\textbf{Rep. Penalty}: 1.2 \\
\textbf{Max Tokens}: 512 \\
\textbf{System Prompt:} \textit{You are an oracle with access to a scene graph about a real-world environment. You are tasked with generating step-by-step instructions for a specific scenario based on the information in the scene graph. The scene graph contains information about the location, setting, objects, their relations, and attributes. However, this scene graph is not necessarily complete. Your goal is to provide detailed instructions that a human or a humanoid can follow to complete the task described in the scenario. For example, if the scenario is "Person A wants to make coffee" and the scene graph shows a kitchen with various objects, the instructions should include steps which Person A needs to take, starting from their current location, to make coffee. You may also be asked for clarifications or additional information by the user if they feel the instructions are unclear or incomplete. Your instructions should be detailed enough for the user to effectively complete the task. Only include steps that are necessary for completing the task and avoid unnecessary details. Beyond making reasonable assumptions about the 3D scene to make up for the potentially incomplete scene graph, try to avoid giving any instructions contrary to the scene. The user does not know that you are an AI oracle and will assume that you are a human providing instructions, and you should act accordingly. Never mention the scene graph or any other AI-related information in your responses, and do not let the user know that you are a language model. Always respond with a single JSON object with instruction index as key and instruction as value. For example, {"1": "Go to the kitchen", "2": "Turn on the coffee machine", "3": "Wait for the coffee to brew"}. Do not output anything else in one message.}

\subsection{Summarizer}

\textbf{Role:} Consolidates the dialogue between Humanoid and Oracle into a refined set of steps, ensuring clarity, coherence, and logical sequence. Configured with a low temperature (0.1) to limit creativity and promote factual and concise summaries, eliminating unnecessary details. \\
\textbf{Model}: llama3-8b-8192 \\
\textbf{Temp}: 0.1 \\
\textbf{Rep. Penalty}: 1.2 \\
\textbf{Max Tokens}: 1024 \\
\textbf{System Prompt:} \textit{You are a Summarizer AI. Given a conversation between different individuals about instructions for completing a task, your goal is to summarize the conversation and provide a concise set of instructions for completing the task. The conversation may contain multiple rounds of questions and answers between the individuals. Your job is to identify the key steps and instructions mentioned in the conversation and provide a summary of the instructions that captures the essence of the conversation. Your summary should be concise and easy to understand, and should include all the necessary steps for completing the task. You should also ensure that the summary is coherent and logically structured, and that it captures all the important details mentioned in the conversation. Your summary should be detailed enough for a human or a humanoid to follow and complete the task described in the conversation. You should not include any irrelevant or unnecessary information in your summary, and should focus only on the key steps and instructions provided in the conversation. Your goal is to provide a clear and concise summary of steps that are required for completing the required task. Do not include any information about the conversation or the individuals involved in your summary, and do not add any additional information beyond the instructions provided in the conversation. Your response should be a single JSON object containing the step-by-step instructions for completing the task, with the instruction index as the key and the instruction as the value. For example, {'1': 'Go to the kitchen', '2': 'Turn on the coffee machine', '3': 'Wait for the coffee to brew'}. Do not output anything else.}

\subsection{Towards Improved Data Quality: A Four-LLM System}

The current multi-agent setup (Humanoid, Oracle, and Summarizer) could be enhanced by introducing a Reviewer agent. The Reviewer would validate Oracle-generated instructions against the complete scene graph ($\mathcal{G}$) to ensure relevance, coherence, and alignment with the scenario-specific graph ($\mathcal{G}_S$). This iterative feedback loop would refine instructions before they are finalized by the Summarizer.

\begin{figure}[ht]
    \centering
    \includegraphics[width=0.85\linewidth]{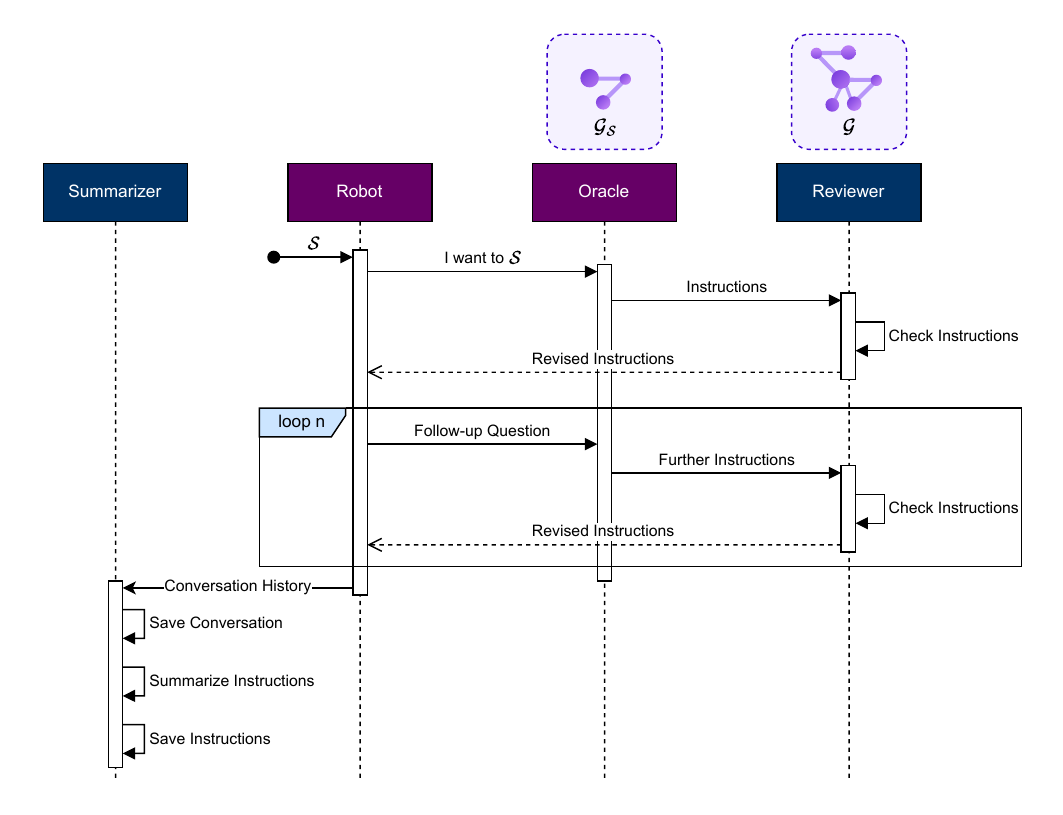}
    \caption{\textbf{Proposed Four-Agent Instruction Generation Pipeline.}  This extension adds a Reviewer agent to the existing system (Humanoid, Oracle, and Summarizer). The Reviewer validates Oracle-generated instructions against the complete scene graph ($\mathcal{G}$), ensuring accuracy, relevance, and coherence before the Summarizer finalizes the instructions. This iterative process aims to further improve the quality and reliability of situational guidance.}
    \label{fig:instructions-generation}
\end{figure}

As shown in Figure~\ref{fig:instructions-generation}, we expect this addition to improve instruction accuracy, reduce errors, and support more complex scenarios. Future work could evaluate the computational trade-offs and potential quality gains of this four-agent architecture, particularly for applications requiring high precision.

\end{document}